Running-head: Code-switching in Human-Machine Dialogs

Strategies of Code-switching in Human-Machine Dialogs

Dean Geckt[1], Melinda Fricke[2], and Shuly Wintner[1]
[1]Department of Computer Science, University of Haifa
[2]Department of Linguistics, University of Pittsburgh

Dean Geckt is now at the Edmond and Lily Safra Center for Brain Sciences, Hebrew University of Jerusalem.

*Acknowledgments: We are immensely grateful to Juan Berríos and Angela Swain for their extensive support in Spanish, including the compilation of the dictionaries, preparation of the maps, and writing the game instructions and the welcome messages of our bot. We are grateful to Yulia Tsvetkov for fruitful discussions and useful comments. We also thank the anonymous reviewers for comments and suggestions that improved the manuscript. This research was supported by Grant No. 2019785 from the United States-Israel Binational Science Foundation (BSF) and by Grant No. 2007656 from the United States National Science Foundation (NSF).

Address for correspondence:

Melinda Fricke
Department of Linguistics
University of Pittsburgh
2822 Cathedral of Learning
Pittsburgh, PA, USA 15260
melinda.fricke@pitt.edu

**Abstract**

Most people are multilingual, and most multilinguals code-switch, yet the characteristics of code-switched language are not fully understood. We developed a chatbot capable of completing a Map Task with human participants using code-switched Spanish and English. In two experiments, we prompted the bot to code-switch according to different strategies, examining (1) the feasibility of such experiments for investigating bilingual language use, and (2) whether participants would be sensitive to variations in discourse and grammatical patterns. Participants generally enjoyed code-switching with our bot as long as it produced predictable code-switching behavior; when code-switching was random or ungrammatical (as when producing unattested incongruent mixed-language noun phrases, such as *'la fork'*), participants enjoyed the task less and were less successful at completing it. These results underscore the potential downsides of deploying insufficiently developed multilingual language technology, while also illustrating the promise of such technology for conducting research on bilingual language use.

## Strategies of Code-switching in Human-Machine Dialogs

Code-switching (CS), in which individuals seamlessly alternate between two or more languages or varieties in conversation, has gained prominence in recent years as researchers seek to understand the social, linguistic, and psycholinguistic factors that shape CS patterns. CS is both ubiquitous in bilingual interaction and widely known to be subject to a variety of conditioning factors, and yet it is still far from being fully understood. Furthermore, CS in human–*machine* interactions has only recently begun to be investigated. Contemporary dialog systems are almost exclusively monolingual, although many bilinguals choose to code-switch when given the opportunity. Understanding whether and how to enhance CS capabilities in dialog systems is crucial for enabling more fluid interactions between humans and machines, and ideally such enhancements can be developed in such a way as to accommodate human conversational preferences and facilitate human behavioral responses.

To this end, the current study addresses fundamental questions regarding CS strategies in human-machine dialog. First, we investigate the mechanisms through which conversational agents can generate more human-like code-switching. Second, we hypothesize that code-switching strategies based on previously attested grammatical patterns correlate with improved task performance and participant satisfaction. Indeed, we find that participants' satisfaction and performance are greatest when conversational agents engage in predictable code-switching strategies (as opposed to random behavior).

This paper makes several contributions. First, we describe the development of an end-to-end system implementing a Map Task, an experimental task widely used for eliciting spontaneous yet experimentally controlled dialogs for the purpose of linguistic analysis (e.g., Anderson et al., 1984, 1991). The online platform we developed is easily expandable and allows for experimentation with various CS strategies. Second, we present a new English-Spanish (alternational and insertional) CS dataset reflecting human-machine interactions collected using the platform. The dataset consists of two separate experiments, each examining different CS strategies. We used a large language model (GPT-4) to carry out the dialog task, and manipulated its responses according to predetermined CS strategies. This resulted in highly fluent bilingual language on the part of the machine, but with deliberately varying degrees of naturalness in its code-switching. Third, we report the novel finding that participants responded positively to more predictable CS strategies, as evidenced by both increased enjoyment and greater objectively measured task success. Most notably, when a code-switching-specific grammatical pattern that is widely expected to be ungrammatical was intentionally introduced, it resulted in a significant decline in overall performance. This has important implications for what chat systems need to strive for. Users will become frustrated and make mistakes if such systems produce grammatically unexpected language. All the resources produced in this work, including the dataset of dialogs and the code for the system, are publicly available.

## 1. Background

Code-switching, the act of shifting from one language to another, has been studied in linguistics and sociolinguistics for decades (Auer, 1998; Clyne, 2003; Gardner-Chloros & Weston, 2015; Gumperz, 1982; Milroy et al., 1995; Muysken, 2000; Sankoff & Poplack, 1981), and more recently, psycholinguists have begun to explore the insights it can provide into language processing (e.g., Beatty-Martínez, Navarro-Torres, & Dussias, 2020). One of the central insights from linguistic research on CS is that it is *non-random*: qualitative linguistic analyses (Bhatt, 1997) and quantitative corpus studies (Pfaff, 1979) alike support the generalization that certain types of switches occur frequently, while others do not. Myriad factors are known to affect the probability and grammatical form of switched elements (see Bellamy & Parafita Couto, 2022; Deuchar, 2020; and Schwieter & Festman, 2023, chapter 5, for relevant reviews), and researchers have taken a variety of approaches to explaining such tendencies (see Aaron, 2015; Poplack, 1980; and Torres Cacoullos & Travis, 2018 for examples of variationist approaches; Balam, Parafita Couto, & Stadthagen-González, 2020; Beatty-Martínez & Dussias, 2017; and Kootstra, Dijkstra, & van Hell, 2020 for experimental investigations; Bhatt, 1997; and Goldrick, Putnam, & Schwarz, 2016 for constraint-based approaches; and MacSwan, 2014 and Myers-Scotton, 2002 for approaches centering grammatical factors).

As an example relevant to our work, consider switches that occur between the determiner and the noun (Bellamy & Parafita Couto, 2022; Clegg, 2006; Myers-Scotton & Gross, 2002; Otheguy & Lapidus, 2003; Pfaff, 1979; Valdés Kroff, 2016). Spanish normally has gender agreement here, as in *'el tenedor* (the fork, masculine) but *'la cuchara'* (the spoon, feminine). When an English noun is inserted into a Spanish grammatical frame, the masculine determiner *'el'* is often used irrespective of the English noun's Spanish translation equivalent (e.g., *'el spoon'* is common, rather than *'la spoon'*). By contrast, the feminine determiner *'la'* is used infrequently and only in cases where the English noun would be feminine in Spanish (e.g., *'la spoon'* is possible but uncommon, while *'la fork'* is not possible). It should be noted that while this pattern has been documented in numerous Spanish-English contexts, the grammatical form of mixed-language noun phrases shows variation due to a number of factors (Beatty-Martínez & Dussias, 2019; Bellamy & Parafita Couto, 2022), and more work is needed to understand the role of community-specific norms in CS (Balam et al., 2020; Deuchar, 2020). Nonetheless, the current study takes advantage of this expected asymmetry to ask whether human participants' performance in a dialog-based game shows sensitivities to grammaticality unexpected CS patterns.

With the rise of social media platforms, written CS (Sebba et al., 2012) has also become a pervasive communication style (Rijhwani et al., 2017). The spoken language domain is not directly comparable to the written one, and findings on CS in written discourse differ somewhat from those in speech (Chan, 2009; Gardner-Chloros & Weston, 2015; McClure, 2001). Our work focuses on written code-switching, and more specifically, on interactions between humans and computational agents (bots). Our primary goal is to improve the quality of bilingual language that is generated by such agents.

Recent works that use neural language models to generate code-switched text have explored various approaches. For instance, Chang et al. (2018) proposed data augmentation techniques to enrich code-switching corpora, while Samanta et al. (2019) employed an autoencoder architecture for generating an extensive corpus of code-switched text. Rizvi et al. (2021) developed a toolkit that generates one code-switched sentence from two monolingual parallel sentences in different languages. However, these methods were not designed for text generation in dialogs. Despite the success of neural language models in general text generation, applications which require CS text generation (e.g., Ahn et al., 2020; Parekh et al., 2020) sometimes prefer rule-based generation methods, relying on contemporary machine translation for translating the utterance's parts.

The work of Ahn et al. (2020) stands out as the first to introduce code-switching into dialog systems. They collected and released *Common Amigos*, a corpus of 587 human–computer mixed Spanish and English text conversations between their dialog system and human participants, facilitating studies of human preferences in written code-switching. *Common Amigos* is an extension of the *Common Friends* game (He et al., 2017), in which two players are each given a list of friends with certain characteristics (e.g., residence, field of study, etc.), and the purpose is to find the one friend they have in common by asking each other questions (e.g., "Do you have a friend who studies geography?"). Ahn et al. examined two main code-switching strategies, *insertional* and *alternational* (Muysken, 2000). The former refers to insertion of elements from one language into the "frame" of the other (Myers-Scotton, 1993, 2002) in a way that generally respects the morphological and syntactic rules of the frame language, while the latter involves alternation of longer, more syntactically complex utterances. Parekh et al. (2020) extended the work of Ahn et al. to an additional language pair, Hindi-English, using the same Common Friends task.

Our work draws inspiration from Ahn et al. (2020) and Parekh et al. (2020), but we employ a more realistic and diversified conversational task, namely the *Map Task* (Anderson et al., 1984, 1991; Thompson et al., 1993; Section 2.1). This is a commonly used game in (psycho)linguistics, where it has been employed to acquire natural (i.e., human–human) dialogs that are simultaneously conversational and spontaneous in nature, but also semi-controlled, experimentally (for a review, see Berríos, Swain, & Fricke, 2023). This task is expected to produce more diverse and complex linguistic interactions (than Common Friends), while still being confined to a restricted vocabulary, and to yield generalizable results that are less domain-dependent.

Like Ahn et al. (2020), we also explore alternational and insertional CS. For the latter, we focus on a particular sub-type of insertional CS, known as *mixed noun phrases*. As noted above, previous research on noun phrases in code-switched English and Spanish has documented an asymmetry in grammatical gender assignment, with a preference for masculine determiners regardless of the Spanish translation equivalent of the English noun. Beatty-Martínez and Dussias (2017) used ERPs to compare processing involving this asymmetry in Spanish-English bilinguals who CS frequently versus those who do not, and found that code-switchers exhibited a heightened sensitivity to (masculine) incongruency manipulations of determiner-noun switches (e.g., *la fork*), manifesting as apparent difficulty in lexical integration. Our study likewise investigates whether human participants are sensitive to gender incongruency manipulations, asking whether this affects their performance in a much more "macro-level" task. Through this investigation we address the broader question of how important it is for dialog systems to "get it right" (i.e., to avoid grammatically unacceptable patterns) when attempting to code-switch.

We begin with a relatively broad investigation designed to test out our methodology and determine whether our task can detect differences in human participants' sensitivity to different CS patterns. Experiment 1 addresses the phenomenon of adaptation between partners during a conversation, known in the literature as *entrainment*, *accommodation*, *adaptation* or *alignment* (Nenkova et al., 2008). Monolingual and multilingual speakers alike are known to seamlessly adjust their communication style to their interlocutors (Bell, 1984; Fricke & Kootstra, 2016; Kootstra et al., 2010, 2012; Pickering & Garrod, 2004), and when interlocutors share more than one language, they nearly inevitably engage in CS. Parekh et al. (2020) examined user accommodation in CS human–machine dialogs and found that humans engage in CS significantly more when the agent code-switches compared to when it converses monolingually. In Experiment 1, our agent generates CS language using different accommodation patterns. We find that our task is able to detect participants' general preference for rule-governed (i.e., non-random) CS. In Experiment 2, we follow up on this finding in more detail, asking whether our task also detects a participant preference for grammatically attested (relative to unattested) CS patterns.

**2. Methods**

We developed an end-to-end online system for the Map Task game, which is currently deployed on the Google Cloud platform, utilizing a server-client architecture. The client component is implemented as a website tailored for human participants, while the server side incorporates the chat-bot implementation necessary for facilitating interactions. All sources and supplementary materials are available at https://github.com/HaifaCLG/MapTask.

## 2.1 The task

The Map Task presents maps containing a set of landmarks both to an *instructor* (tasked with communicating the path depicted from a start point to an end point, using language only) and a *navigator* (tasked with faithfully reproducing the instructor's path on their own map). The dialog produced during this interaction is the main object of research.

In our setup, each experimental session comprises four Map Task games (Figure 1), each featuring a unique map. The start point (marked with 'X') is always at the top of the map, and the end point is marked by a checkmark. See Berríos et al. (2023) for additional information on the development of these maps. Our human participants assume the roles of both instructor and navigator in a predetermined sequence: instructor, navigator, instructor, and navigator.

<Figure 1 about here.>

## 2.2 Participants

Participants were recruited using Prolific and then directed to our Map Task website. To help ensure both English and Spanish proficiency, we used Prolific's screening procedures to invite only participants who were self-reported fluent speakers of Spanish residing in the U.S., and we presented our Prolific advertisement in Spanish.

Participants were informed that they would be working with a partner to either give or receive directions on a map, and that they could do this using a dedicated chat box. To encourage them to be more talkative, we informed them that this was a research study about how people use language to communicate information, and that their performance in the study may affect their eligibility for a financial bonus. The experimental session lasted about 40 minutes in total, and participants were paid the equivalent of £6, with a £1 bonus as long as they appeared to put in a good faith effort. The full task instructions are available in Appendix A (see supplementary materials).

We recruited 50 participants per experimental condition, with the total number of participants retained in each analysis given in Tables 2 and 4. Participants who did not perform the navigation task (i.e., whose path was empty) or who did not engage in conversation were excluded from the analyses. Additionally, participants who did not specify either English or Spanish as a native (or second) language in our end-of-session language questionnaire were excluded. However, we did not perform any further filtering based on the questionnaire responses due to the pre-screening provided by Prolific, and because we suspected that some participants may have skipped through the questionnaire rather quickly.

Table 1 provides self-reported demographic and language information for participants in Experiments 1 and 2. While a goal of recruiting participants currently residing in the U.S. was to reduce variability in the data somewhat, we acknowledge that our analyses lump together speakers from a wide variety of language communities and linguistic backgrounds. While most (~75%) participants were born in the U.S. and considered Spanish their second or less proficient language (~69%), participants lived in a variety of U.S. states and reported various home language environments (e.g., ~50% reported English as their only native language, versus ~25% reporting Spanish only and ~25% reporting English and Spanish). Table 1 is intended primarily to document potentially relevant variation among our participants, and we hope that future research will employ the methods we develop here to better understand how such variation impacts CS patterns.

<Table 1 about here.>

## 2.3 System architecture

**Client-side.** This is a web application that allows participants to play both roles of the Map Task game, chat via a dedicated interface, and respond to questionnaires (see Figure 2). When playing as the navigator, participants build their path using an avatar on the map that can be moved with the mouse and keyboard. More detail on the client is given in Appendix A.

<Figure 2 about here.>

**Server-side.** This is a Python application that implements the conversational agent (*chatbot*, or just *bot*, below) and records the data for each experiment. The bot contains the implementation of the code-switching strategies described below, and is based on GPT-4 (OpenAI et al., 2024). We use different prompts to handle the roles of instructor and navigator, and to transmit knowledge about each map. The code flow of the bot consists of two main stages:

**1) Generate a new response.** The participant's most recent response is passed along with the chat history to GPT to generate a suitable response based on the bot's role in the game.

**2) Apply code-switching strategy.** At this stage, we (potentially) manipulate the response of GPT, based on the CS strategy.

More detail on the server is given in Appendix B.

## 2.4 Overview of experimental setup

The setup for Experiments 1 and 2 is globally similar. We initially provide a set of instructions (a prompt) to GPT-4 and then in the course of the experiment, we potentially manipulate its output according to a particular CS strategy. Each strategy corresponds to a single condition, such that CS strategies in both experiments constitute a between-groups manipulation.

In Experiment 1 (our "alternational" experiment), we do not instruct GPT-4 to communicate in any particular language, but rather use machine translation to translate its *entire* output from one language to the other, depending on the strategy. In Experiment 2, by contrast, GPT-4 is explicitly instructed to communicate in Spanish, and the messages presented to participants are in mixed Spanish-English (i.e., following an "insertional" strategy), with their precise grammatical form depending on the condition. In both versions of the experiment, the chat begins with one of a small set of predefined welcome messages (see Appendix B), in order to establish a social foundation. The prompts to GPT-4 are written in English, and its initial output in fact depends on the participant's language usage.

## 2.5 Computational resources

We implemented several language processing modules to modify the bot's generated utterances in line with our CS strategies.

**Language identification (LID).** We classify each sentence into four categories: English, Spanish, Mixed, and None. This is done using a pre-trained transformer-based model (Vaswani et al., 2017) at the token level: each token (i.e., word) is classified as either *English* or *Spanish*. Then, the entire utterance is classified as *Mixed* if it includes at least one token from each language. If all tokens belong to a single language, the utterance is labeled accordingly. The label *None* is given to short (typically, one-word) utterances such as *'ok'* or *'no'*, where language identification is ambiguous.

**Noun extractor.** As one of our experiments focuses on code-switching around noun phrases (NPs), we use SpaCy (Honnibal & Montani, 2017) to parse the sentences. We then identify *simple noun phrases* by focusing on nouns: we use the dependency structure to identify the determiner of each noun. Complex noun phrases, such as those containing adjectives, possessives or relative clauses, are disregarded during this process, such that our experimental manipulations affect only simple NPs.

**Translation.** For some strategies (in Experiment 1), we use Google's translation services to translate the entire utterance.

**Spanish-to-English noun dictionary.** We constructed a dictionary that contains the expected Spanish-to-English translations of 135 objects found on the maps, including common synonyms for some objects. It is used to translate a single noun identified by the noun extractor component, in order to generate mixed NPs.

**English noun gender dictionary.** We constructed a dictionary that lists the Spanish translation equivalents of English nouns listed in the previous dictionary, along with their Spanish gender. For example, an expected translation of the noun *'rock'* to Spanish is *'roca',* which is feminine, hence the item *'rock–feminine'* is included in this dictionary. The dictionary includes 109 nouns, fewer than the previous dictionary since some Spanish synonyms are mapped onto the same English noun. In cases where possible Spanish synonyms have different genders (for example, *'papagayo'* is masculine whereas *'guacamaya'* is feminine, but both are translated to *'parrot'*), the gender is listed in this dictionary as ambiguous.

**Masculine-to-feminine mapping of Spanish determiners.** These are actually two dictionaries containing mappings from masculine to feminine and vice versa, adhering to Spanish grammatical rules. This allows any alteration of Spanish determiners to select the contextually appropriate forms.

All dictionaries were curated by two Spanish-English bilinguals.

## 2.6 Dependent measures

Some of the dependent measures of interest are computed from the dialog, while others are reported by the experiment participants via questionnaires. A full description of the questionnaires is given in Appendix C.

**Task performance.** We focus on two quantitative measures:

1) *Time taken to complete a single game*, where each map has a time limit of seven minutes.

2) *Route similarity.* The participant's route is compared to the target route and given a numeric score using the *dynamic time warping* algorithm (Salvador & Chan, 2007), which finds optimal temporal alignment between two sequences; we use Manhattan distance as the basic distance measure, so all distances are integer values, but we normalize the resulting score by dividing it by the length of the target path.

**Participant satisfaction.** Each participant was asked to respond to the following questions using a scale from 0 to 100, following the completion of the experimental task:

**Task Enjoy.** "How much did you enjoy the task?"

**Task Success**. "How successful do you think you were at completing the task?"

**Difficult Comm.** "How difficult was it to communicate with your partner?"

**Difficult Ins.** "How difficult was it to understand your partner's instructions?"

 In addition, we also analyze two metrics specific to Experiment 1:

**Inter-sentential CS.** This metric tracks language shifts within the dialog for the human participant. A switch is tallied when the language of sentence $i$ differs from that of sentence $i+1$, where neither is mixed language.

**Inter-sentential entrainment.** This metric counts the number of utterances in which the participant's language is the same as the bot's language in the preceding utterance. We consider mixed language sentences as wildcards, i.e., equal to either Spanish or English.

We use **task success** as a general term to refer to both *task performance* and *participant satisfaction*, either jointly or individually, and in the analyses that follow, we ask whether task success, inter-sentential CS, or inter-sentential entrainment vary as a function of CS strategy.

## 3. Experiment 1: Alternational code-switching

### 3.1 Research questions

In this experiment, we examine the relationship between discourse-level characteristics of CS and task success. We hypothesize that:

1) Participants may experience lower task success when the bot's language choice is anti-cooperative and/or unnatural.

2) Greater entrainment (on the part of the participant and/or the bot) may be associated with greater task success (Reitter & Moore, 2014).

### 3.2 Experimental conditions

In Experiment 1, we investigate five CS strategies:

**Baseline.** Display the bot's response without translation or modification. This strategy serves as a control.

**Alignment.** If the last human utterance was in one language (specifically, not mixed), respond in the same language. This strategy builds on evidence that bilinguals tend to accommodate their interlocutors in general (Trofimovich & Kennedy, 2014), and adapt to their CS style in particular (Parekh et al., 2020), with the possibility of converging to a totally unilingual dialog.

**Adversarial.** If the last human utterance was in one language, respond in the *other* language. This (presumably unnatural) strategy is expected to generate the largest amount of intersentential CS per dialog.

**Random.** Switch the language of GPT's response in a randomly determined 50% of cases. In all other cases, do nothing. If the original response is in mixed English-Spanish, retain it as is. The motivation behind this strategy is to produce a dialog in which language switching is present, but follows no obvious pattern.

**Short Context.** If the last $k$ utterances of the bot were in the same language (bot utterances that mix English and Spanish are not counted), switch to the other language. We set $k$ to be 3. The first few ($< k$) bot utterances are retained intact. The motivation for this strategy is to generate language switches at a constant rate.

Participant characteristics across the two experiments are summarized in Table 1 above, while Table 2 gives the breakdown across the experimental conditions in Experiment 1 only.

<Table 2 about here.>

### 3.3 Analysis and Results

Table 3 gives detailed descriptive statistics characterizing the dialogs collected in both Experiments 1 and 2.

**Task performance and participant satisfaction.** The data were analyzed using ANOVA, with post hoc pairwise $t$-tests to follow up on significant effects, using the multivariate $t$ adjustment for multiple comparisons as implemented in the *emmeans* package (Lenth, 2024), as appropriate. Dependent variables were transformed to improve normality as warranted; this affected the analyses of *Game Time* (arcsin transformed), *Route Distance* (log transformed), *Task Enjoy*, *Task Success*, *Difficult Comm*, and *Difficult Ins* (all arcsin transformed).

**Game Time.** There was a main effect of role only ($F(1, 426) = 146.6, p < .001$); completion times were slower for navigators compared to instructors (mean times of 326 s vs. 240 s; $t(426) = 12.1, p < .0001$).

**Route Distance.** Route Distance did not differ significantly according to strategy, human role, or their interaction (all $F$s < 1.6, $p$s > .22).

**Task Enjoy.** There was a main effect of strategy ($F(4, 426) = 3.1$, $p = .02$) and a main effect of role ($F(1, 426) = 7.4$, $p = .007$). Only the *random* condition differed from the *baseline* condition (mean enjoyment ratings of 70.0 vs. 83.6; $t(426) = -3.4$, $p = .003$). Across all conditions, participants enjoyed being instructors more than navigators (mean ratings of 81 vs. 75, respectively; $t(426) = 2.7$, $p = .007$).

**Task Success.** The analysis of self-reported success revealed a main effect of role only ($F(1, 426) = 146.6$, $p < .001$); participants self-reported greater success when they were instructors compared to when they were navigators (74 vs. 70; $t(426) = 2.2$, $p = .03$).

**Difficult Comm.** There was a main effect of strategy ($F(4, 426) = 3.7$, $p = .006$) and a main effect of role ($F(1, 426) = 7.3$, $p = .007$). Only the *random* condition differed from the *baseline* condition (mean difficulty ratings of 43.5 vs. 29.8; $t(426) = 3.3$, $p = .004$). Across all strategies, participants found communication more difficult when they were navigators compared to when they were instructors (38 vs. 31; $t(426) = 2.7$, $p = .007$).

**Difficult Ins.** There was a main effect of role only ($F(1, 426) = 5.0$, $p = .03$); participants found understanding more difficult when they were navigators compared to when they were instructors (40 vs. 32; $t(426) = -3.0$, $p = .003$).

**Statistical analysis of CS behavior.** To better understand participants' linguistic behavior and how it might have affected their experience of the task, we also analyzed the inter-sentential entrainment and inter-sentential CS metrics. For inter-sentential CS, there was a main effect of strategy ($F(4, 426) = 43.1$, $p < .0001$). As compared to the *baseline* (mean = 6.0%), all of the strategies showed a greater proportion of utterances with inter-sentential switches, except for the *alignment* condition (mean = 3.1%), which did not differ from the *baseline* ($t(426) = -1.2$, $p = .59$). For *adversarial*, the mean was 32.9% ($t(426) = 10.3$, $p < .0001$); for *random*, the mean was 20.6% ($t(426) = 6.0$, $p < .0001$); and for *short-context*, the mean was 18.2% ($t(426) = 5.2$, $p < .0001$).

The analysis of inter-sentential entrainment returned main effects of strategy ($F(4, 426) = 20.9$, $p < .0001$) and role ($F(1, 426) = 20.6$, $p < .0001$). All strategies showed less entrainment compared to the *baseline* condition (mean = 84.4%) except for the *alignment* condition (mean = 89.0%), which did not differ from *baseline* ($t(426) = 1.5$, $p = .4$). For *adversarial*, the mean was 66.5% ($t(426) = 5.7$, $p < .0001$); for *random*, the mean was 73.4% ($t(426) = 3.5$, $p = .002$); and for *short-context*, the mean was 67.4% ($t(426) = 5.5$, $p < .0001$). Across all conditions, instructors demonstrated more entrainment than navigators (0.80 vs. 0.72; $t(426) = 4.5$, $p < .0001$).

### 3.4 Experiment 1 Discussion

Most differences across the alternational experiments involved the user's role in the dialog. In general, participants were less successful as navigators than as instructors, and perceived nagivating to be more difficult than instructing. However, these role effects were consistent across the different CS strategies.

In terms of differences in task success across CS strategies, only the *random* strategy differed significantly from the *baseline.* Introducing random code-switching behavior caused participants to enjoy the task less and rate the communication task as more difficult. We assume that language choice is less predictable under this condition, and participants are often surprised, leading to difficulties in understanding their interlocutor. This partially corroborates our hypothesis that anti-cooperative and/or unnatural language choices on the part of the bot may lead to lower task success. To our surprise, however, the *adversarial* strategy did not show a similar effect. It is possible that participants perceived the strategy that we imagined as adversarial to actually be somewhat accommodating. To illustrate from the participant's point of view, in this condition if the bot is speaking only Spanish and the user is speaking English, then as soon as the user switches to Spanish, the bot will switch to English. This could be perceived as a cooperative behavior, i.e., "If I switch to Spanish for you, it's nice that you switch to English for me." However, this is only conjecture, and further investigation is needed to test this proposal.

In terms of participants' own language use, the *baseline* strategy was not characterized by many inter-sentential switches, meaning the dialogs were predominantly unilingual, while the *adversarial* strategy had the most inter-sentential switches, indicating a more bilingual dialog. Additionally, due to being largely unilingual, *baseline* and *alignment* were the CS strategies that yielded the highest inter-sentential entrainment, i.e., participants aligned to the bot's (single) language the most under these two conditions, while the other strategies showed less entrainment. However, task success was only significantly lower than the *baseline* under the *random* strategy, so we did not find strong evidence supporting our second hypothesis.

Interestingly, quite a few utterances (around 10%, averaging across all conditions) were classified as mixed, and some of these contained noun phrase (NP) switches, where a language switch occurs between the determiner and the noun, as in '*el fork'*. Importantly, we did not explicitly instruct participants to code-switch, nor was the bot instructed or manipulated to produce such switches. The fact that participants still produced NP switches under these conditions, and that their production preferences generally aligned with previously observed tendencies in terms of which types of NPs switches occur most frequently in bilingual communities, is an interesting finding of this experiment. The bot's inter-sentential code-switching seems to have motivated the participants to code-switch themselves, and their CS patterns naturally contained insertional (mixed NP) switches. While this was not the subject of Experiment 1, we decided to explicitly investigate this phenomenon in Experiment 2.

## 4. Experiment 2: Insertional code-switching

### 4.1 Research questions

In this experiment we explore the behavior of human participants when our conversational bot produces insertional noun phrase (NP) code-switches. We hypothesize that:

1) Participants may experience lower task success when the bot produces code-switches that have been associated with processing difficulty in previous work (i.e., the masculine incongruent condition, *'la fork'*).

2) Participants will generally tend towards entrainment, such that their grammatical patterns tend to align with those of the bot.

### 4.2 Experimental conditions

This experiment includes four CS strategies that the chatbot (instructed in its prompt to communicate exclusively in Spanish) employs as it interacts with the human participant. To implement these strategies online we use the language identification and dictionaries described in Section 2.5 to translate nouns and change the gender of determiners. Below are the four strategies and our expectations of each:

**Baseline.** Display the bot's response without translation or modification. We expect this condition to result in overwhelmingly Spanish-language conversation.

**Congruent.** Translate nouns from Spanish to English, and ensure that all Spanish determiners match the original noun's gender. For example, an initially generated NP such as *'el tenedor'* would be translated to *'el fork'*, and *'la cuchara'* would be translated to *'la spoon'*. We expect this condition to be perceived as relatively natural, and hence to not have a negative effect on task success.

**Feminine Incongruent.** For feminine nouns only, translate the noun and switch the determiner's gender. For example, an initially generated NP such as *'la cuchara'* would be translated to *'el spoon'*. We expect this condition to be broadly acceptable, based on previous research.

**Masculine Incongruent.** For masculine nouns only, translate the noun and switch the determiner's gender. For example, an initially generated NP such as *'el tenedor'* would be translated to *'la fork'*. Based on previous research, this is the most unnatural condition, and we expect it (1) not to be present at all in the human utterances, and (2) to have a detrimental effect on task success.

These strategies are implemented as follows: For every generated response, we utilize the *noun extractor* component to identify all simple noun phrases. Then, each Spanish noun is translated using our Spanish-to-English noun dictionary. Finally, to switch the Spanish determiner's gender we use our masculine-to-feminine mapping.

Table 4 gives the breakdown of participant characteristics across the experimental conditions in Experiment 2 only.

<Table 4 about here.>

## 4.4 Dependent measures

To evaluate task performance we use all the measures defined in Section 2.6, plus counts of the following four types of mixed NPs. We focus on the quantification of simple NP switches, where the matrix language of the sentence is Spanish, and an English NP is inserted into it. Below, 'noun' refers to the Spanish translation equivalent of the English noun that occurred in the utterance (according to our dictionary), and 'determiner' is the Spanish determiner that precedes the English noun. In cases where the noun's gender cannot be determined (because there are multiple straightforward translation equivalents and their genders differ), the NPs are categorized as ambiguous and excluded from analysis.

**Congruent masculine.** Both the noun and the determiner are masculine.

**Congruent feminine.** Both the noun and determiner are feminine.

**Incongruent masculine.** The noun is masculine while the determiner is feminine.

**Incongruent feminine.** The noun is feminine while the determiner is masculine.

Table 3 gives descriptive statistics for the participants' performance (i.e., ignoring the bot's utterances and averaging across the navigation and instruction tasks).

## 4.5 Analysis and Results

**Task performance and participant satisfaction.** The data were analyzed as in Experiment 1, with dependent variables again transformed as warranted to improve normality; this affected *Game Time* (arcsin transformed), *Route Distance* (log transformed), *Task Enjoy*, *Task Success*, *Difficult Comm*, and *Difficult Ins* (all arcsin transformed). In Experiment 2, rather than collecting participants' evaluations after each map, we collected self-evaluations a single time, after the final map. Consequently, we do not test for effects of participant role on self-reported data in Experiment 2.

**Game Time.** There was a main effect of role ($F(1, 334) = 71.8$, $p < .001$), and an interaction between role and strategy ($F(3, 334) = 3.2$, $p = .02$). The main effect of strategy was not significant ($F(3, 334) = 1.2$, $p = .3$). Follow-up *t*-tests indicated that averaging over all conditions, navigators completed the task more slowly than instructors (mean times = 338s vs. 261s, respectively; $t(334) = 8.5$, $p < .0001$), and that the interaction between role and strategy was driven by the fact that navigators in all conditions except the *feminine incongruent* condition completed the task more slowly than instructors (all *t*s > 4.5, all *p*s < .0001, except $t(334) = 1.7$, *p* = .3 for the *feminine incongruent* condition). In other words, the slowing effect of being a navigator as compared to an instructor was consistent across all conditions except for the *feminine incongruent* condition, where navigators were able to complete the task particularly quickly. This effect is depicted in Figure 3, and suggests that comprehension in particular may have been easier in the *feminine incongruent* condition as compared to the other conditions.

<Figure 3 about here.>

**Route Distance.** Objectively measured distance scores differed by strategy ($F(3, 334)$ = 3.9, *p* = .009), but not by map ($F(1, 334)$ = 0.8, *p* = .4) or their interaction ($F(3, 334)$ = 0.1, *p* = .9). Follow-up *t*-tests demonstrated that the effect of strategy was driven by the fact that in the *feminine incongruent* condition, distance scores were lower than in the *baseline* (mean distance scores = 2.2 vs. 3.1; $t(334)$ = -2.8, *p* = .02), indicating that objectively measured task success was better in the *feminine incongruent* condition as compared to the *baseline*, and this effect was consistent across maps.

**Task Enjoy.** There was a main effect of strategy ($F(3, 338)$ = 6.0, *p* <.001), such that enjoyment was lower for both the *congruent* ($t(338)$ = -2.7, *p* = .02) and *masculine incongruent* conditions ($t(338)$ = 4.0, *p* < .001) as compared to the *baseline*, with mean enjoyment ratings of 77.9 (congruent), 74.7 (masculine incongruent), and 86.0 (baseline). The *feminine incongruent* condition did not differ from the *baseline* (means of 84.1 vs. 86.0, respectively; $t(338)$ = −1.2, *p* = .5).

**Task Success.** For self-reported success ($F(3, 338)$ = 2.6, *p* = .04), none of the conditions differed significantly from the *baseline* (all *t*s < 1.7, all *p*s > .24). Instead, the main effect of strategy was driven by a difference between the *masculine incongruent* and *feminine incongruent* conditions ($t(338)$ = 2.7, *p* = .006); participants self-reported somewhat less success in the *masculine incongruent* condition than in the *baseline* (means of 67.8 vs. 69.5, respectively), and somewhat more than *baseline* in the *feminine incongruent* condition (75.0 vs. 69.5).

**Difficult Comm.** For difficulty communicating ($F(3, 338) = 5.3$, $p = .001$), only the *masculine incongruent* strategy differed from the *baseline* (mean ratings of 48.6 vs. 37.3, respectively; $t(338) = 2.6$, $p = .03$).

**Difficult Ins.** For difficulty understanding ($F(3, 338) = 5.8$, $p < .001$), only the *feminine incongruent* strategy differed from the *baseline* (means of 25.7 vs. 36.7; $t(338) = −2.7$, $p = .02$).

**CS behavior.** We also analyzed participants' own linguistic production; specifically, we counted the total number of NP switches that were produced across all of the experiments, and we used a chi-squared test to ask whether the relative proportion of these differed across CS strategies. Table 3 shows the result of this counting, with proportions calculated within each column in parentheses (thus asking the question, "within a given condition, what proportion of NPs were of each grammatical type?"). Values for which the absolute *Pearson* residuals from the chi-squared test were greater than 2 (i.e., those that differed significantly from the expected value) are marked with an asterisk. The chi-squared test returned a significant result ($\chi^2(9) = 35.5$, $p < .0001$), which was driven by performance in the *masculine incongruent* condition; this condition yielded fewer feminine incongruent switches and more masculine incongruent switches as compared to the other conditions.

However, this analysis collapses across participants. To determine whether differences in production were generalizable across participants, we counted the number of people who produced each type of NP switch across all experiments. A chi-squared test on these data was marginal ($\chi^2(9) = 16.0$, $p = .07$), indicating that in general, any differences across cell counts may well have been due to chance. However, the Pearson residuals for a single cell were greater than 2.0: the number of participants producing masculine incongruent NPs in the *masculine* *incongruent* condition (n = 5, versus just 1 or 2 in the other conditions), indicating that of the participants who produced masculine incongruent NPs, more were in the *masculine incongruent* condition than would be expected by chance. Along with the tabulation of NP counts above, this indicates that bot behavior in the *masculine incongruent* condition affected participants' language use, though the effect was driven by a small number of participants.

As seen in Table 3, participants produced the smallest number of mixed NPs (n = 17) in the *baseline* strategy, whereas in other strategies the quantities are much higher (n = 58 or more). This is a strong indication of *alignment*: participants seem to have adapted their CS style to the bot, producing many more mixed NPs when the bot did so than when the bot did not. Additionally, the most common NP switches are congruent masculine and incongruent feminine; in both, the determiner is masculine, in line with previous findings indicating an overall preference for masculine determiners (Valdés Kroff , 2016). In contrast, under the *masculine incongruent* condition the number of incongruent feminine switches produced by participants was significantly lower compared to the other conditions. In fact, participants in this condition hardly made any incongruent feminine switches, despite them being widely considered grammatical. This is another indication of alignment; under this condition the bot did not produce any incongruent feminine switches, and therefore neither did the participants.

### 4.6 Experiment 2 Discussion

In terms of path quality and self-report of enjoyment, success, and difficulty communicating with the bot, the most successful strategy was *feminine incongruent*. In addition, participants were generally slower to complete the task when acting as navigators than as instructors, but this effect was greatly reduced in the *feminine incongruent* condition. Taken together, these results suggest that participants may have found it particularly easy to understand the bot's instructions when they contained grammatically expected code-switches, and this ease in language comprehension appears to have made the Map Task itself easier and more enjoyable.

Our analysis of participants' own linguistic production also showed some modest evidence of entrainment to the bot's grammatical patterns. In the *masculine incongruent* condition, participants produced unexpectedly few feminine incongruent switches, while also producing more masculine incongruent switches than in the other conditions. Given that feminine incongruent switches are widely considered grammatical across bilingual communities (e.g., Valdés Kroff, 2016), it is perhaps not surprising that they were produced at a relatively high rate across most conditions. Feminine incongruent switches constituted from 29% to 39% of the total number of NP switches in the *baseline*, *congruent*, and *feminine incongruent* conditions, more than the proportion of feminine congruent switches in each case.

It *is* surprising, however, that participants produced so few feminine incongruent switches in the *masculine incongruent* condition. In this condition, the bot consistently produced ungrammatical mixed NPs such as *la tiger, la tree,* and *la crocodile*, while never producing the more expected feminine incongruent mixed NPs (such as *el giraffe, el snake, el butterfly,* etc.). The combination of the lack of expected constructions and the presence of ungrammatical constructions not only resulted in lower task enjoyment and self-reported success in the *masculine incongruent* condition, and in greater difficulty communicating, it also appears to have affected participants' own willingness to code-switch. Importantly, while feminine incongruent switches in particular were much reduced in the *masculine incongruent* condition, the total number of mixed NPs was considerably lower in this condition as well (only 58, vs. 103 and 110 in the other conditions in which the bot was specifically manipulated to code-switch). One interpretation of this finding is that when a bot code-switches *poorly*, participants are less inclined to code-switch themselves, and they are especially less likely to produce grammatical patterns that they perceive to be colloquial, such as feminine incongruent switches. We were also surprised to find that a handful of participants themselves produced incongruent masculine switches in this condition, given that this construction is widely considered ungrammatical. However, this finding was constrained to a small number of participants, so we do not wish to overstate its importance.

Finally, we were somewhat surprised to find that participants rated their task enjoyment lower in the *congruent* condition as compared to the *baseline*. In this condition, the bot produced roughly twice as many mixed NPs as compared to the two *incongruent* conditions (because all nouns were translated to English rather than only nouns of a particular gender). Future work should examine the question of whether the number and/or regularity of code-switches can help predict participants' enjoyment in human-machine interactions.

## 5. General Discussion

### 5.1 Comparing alternational and insertional CS in our experiments

While our two experiments were built on the same platform and were similar in many ways, future work can benefit from an explicit consideration of the ways in which they differed. Most notably, in Experiment 1 we did not instruct GPT-4 as to which language to use, while in Experiment 2 we told it to communicate in Spanish. This difference is reflected in the percentage of utterances produced by participants in each language (31% Spanish in Experiment 1 vs. 65% Spanish in Experiment 2, averaging across conditions). Second, in Experiment 1 we used machine translation to translate full utterances of GPT-4's output according to our CS strategies, while in Experiment 2, we used online parsing and custom-prepared dictionaries to extract, translate, and manipulate particular noun phrases. These differences are reflected in the number of inter-sentential switches produced by our participants and in the number of mixed NPs in particular, but *not* in the number of mixed-language utterances, which was quite similar across experiments. These findings can be taken as high-level evidence for entrainment, and Experiment 1 in particular also suggests that as long as a bot demonstrates some facility in both of a participant's languages, some non-negligible proportion of the participant's utterances are likely to contain utterance-internal code-switching.

In terms of the amount of language produced, Table 3 shows that the number of utterances and tokens produced in each of our experiments were similar. The proportion of utterances containing both English and Spanish ("Mix") was also similar across experiments, as was participants' self-reported task enjoyment and task success. However, game durations were shorter and path differences were smaller, on average, in our alternational experiment as compared to the insertional one. These differences may have been due to the fact that many of our participants were English-dominant, and the majority language of the insertional experiments was Spanish.

## 5.2 Comparison with previous work

Similar to us, Ahn et al. (2020) and Parekh et al. (2020) compared two main code-switching strategies, *insertional* and *alternational*. They collected and released two datasets, in two different language pairs (Spanish-English and Hindi-English, respectively), using the same *Common Friends* task. In this section we refer to their results, jointly, as the *Common Friends* experiments. While our goals were influenced by the *Common Friends* experiments, our analyses focused more heavily on how different (natural and unnatural) CS strategies affect task success at different levels. Our work takes advantage of state-of-the-art language technology (namely, a far more sophisticated chatbot based on GPT-4), and compared with the Common Friends task, the Map Task is a more realistic and diversified conversational task. Because of the richness of the data generated in the Map Task, we were able to examine a wide variety of metrics, producing a more complete picture of task success than was available in previous work.

Our alternational strategies differ from the *Common Friends* experiments in that we switch the language of the *entire* sentence, while they switched only parts of it. Another difference is that they differentiate in the directionality of the switch, either English inserted into Spanish/Hindi or vice versa, whereas we focus only on English-into-Spanish switches in our insertional experiments, and in our alternational experiments, the direction of the switch varies throughout the dialog in a pattern that depends on the condition. Still, as in their work, our insertional strategies also focus on nouns or noun phrases.

Ahn et al. (2020) also experimented with a parameter of *informality*, operationalized using discourse markers, in order to test the hypothesis that CS is observed more often in informal and casual settings than in formal ones. Inspired by their work, we attempted to create a casual setting by beginning each chat with an informal welcome message, and by prompting our bot to imagine playing a dialog-based game "with a good friend" (see Appendix B). However, we did not include formality as a variable in our experiments.

The results of Ahn et al. (2020) established that 1) the insertional strategy, specifically English inserted into Spanish (the same direction we employed), was the most frequent in the corpus, and the one in which task success was the highest; and 2) humans sometimes adapted to the agent's CS, but their choice of CS strategy primarily depended on their language proficiency. Parekh et al. (2020) showed that participants code-switched significantly more when the agent was code-switching, compared to when the agent was monolingual.

To facilitate further comparisons with *Common Friends*, we computed a binary success metric, as they did: the *Common Friends* researchers considered the percentage of successful games, where a successful game was one in which the common friend was found. We dichotomized success in our Map Task by computing the percentage of games where the goal was reached in less than seven minutes. Table 5 gives a subset of the results of the *Common Friends* experiments (i.e., only those with the highest success rates) along with our results, aggregated to two rows containing the grand averages of our two experiments. In general, longer dialogs were produced in our experiment. This is not surprising since the Map Task requires participants to share more information than *Common Friends*. On the other hand, *Common Friends* conversations had a higher percentage of mixed utterances. This is likely due to a difference in instructions; according to Ahn et al. (2020), their instructions encouraged users to "use English, Spanish, or a mixture of the two" and the title of their task was *'Charlemos en Spanglish!',* "Let's chat in Spanglish!". Our instructions, on the other hand, did not mention mixture of languages, bilingualism, or the word *'Spanglish'* (see Appendix A.1). It is also worth noting that naturally occurring CS corpora tend to exhibit rather low rates of mixed utterances (e.g., the Bangor Miami corpus includes less than 6% utterances with an intrasentential switch; Fricke & Kootstra, 2016).

According to Table 5, in the *Common Friends* experiments, insertional strategies were more "successful" as compared to alternational, while the reverse was true in our experiments. However, our success rate is higher on average (about 85%, aggregated across alternational conditions) compared with *Common Friends* (highest score of 77%). Given the greater task complexity in our experiments, it is notable that our participants were generally more successful at completing their task, and that they generated more language in doing so. Moreover, our participants were spontaneously motivated to code-switch with our chat agent despite any explicit instruction or suggestion to do so on our part, and given the more detailed analysis of our results above, we would also underscore that our experimental setup was able to capture "macro" differences in participant behavior as a function of relatively "micro" differences in grammatical usage. We therefore submit that the Map Task and its implementation in our experiments shows the greatest future promise for researchers wishing to study naturalistic code-switching patterns.

## 6. Summary, Limitations, and Conclusions

We created a novel bilingual dialog system and recruited a large number of participants to converse with our chat agent. These experiments resulted in two human-machine dialog datasets for Spanish-English. To the best of our knowledge, our system is the first of its kind to examine different code-switching strategies using a chatbot based on a large language model, and as a result, rich, linguistically diverse dialogs were produced. All of our data and code are publicly available.

We carefully manipulated the behavior of our bot to explore participants' responses to a variety of code-switching strategies. We found that basic conversation characteristics (such as the number of utterances and tokens, and even the number of mixed-language utterances) were very similar across strategies, and both experiments showed indications of alignment. In our alternational experiment, where the language of entire sentences was switched across utterances, we found that strategies with unpredictable behavior, such as *random*, led to poor task performance. More predictable strategies, such as the *baseline* and even the *adversarial* strategy (which performed better than expected), yielded higher success rates according to some performance metrics.

In our insertional experiment, we replicated previous findings that *feminine incongruent* NP switches were widely used by participants, and were even associated with higher success rates in our Map Task. In contrast, participants very rarely produced *masculine incongruent* switches, and when the bot used them it resulted in poorer task performance. Surprisingly, our *congruent* strategy revealed that higher overall rates of insertional switches produced by the bot negatively impacted task success. More research is needed to better understand this finding.

Previous research (Ahn et al., 2020; Parekh et al., 2020) suggested that insertional strategies were more successful than alternational ones, but only when task success was defined in simple terms. In contrast, our comparison of the strategies, which relied on more elaborate success metrics, showed the opposite. We attribute this difference to the nature of the experiment (the bot in our insertional experiment communicated predominantly in Spanish) combined with our sample of participants (who tended to be English-dominant). Whether insertional or alternational, code-switching strategies that felt more natural to participants yielded higher success rates, while unnatural strategies generally resulted in worse performance.

While our work sheds new light on the nature of code-switching, especially in human-machine dialogs, many questions are left for future work. One question is the impact of the number of switches the agent performs on task success. Is there a number or proportion of switches that leads to negative outcomes (as was the case in our *congruent* strategy)? Moreover, consistent with previous work (e.g., Valdes Kroff, 2016), we observed a clear preference for NP switches in which the determiner was masculine, regardless of the Spanish translation equivalent of the English noun. However, as acknowledged in the Methods section, a potential limitation of the current work is that our participants represented a heterogenous sample of language profiles and speech communities, obscuring the impact of these factors on CS behavior in our experiments. Given research suggesting that community norms may be a significant driver of CS patterns (Balam et al., 2020; see also Beatty-Martínez et al., 2020; Bellamy & Parafita Couto, 2022; Blokzijl, Deuchar, & Parafita Couto, 2017), future work should investigate the impact of these variables (and others) on CS preferences. Our system makes it is possible to quickly collect large amounts of data to help address these questions, and with relatively small adjustments to the architecture, it can also facilitate similar investigations in a variety of other language pairings.

In fact, many interesting directions for future research are related to the technology we used in this work. State-of-the-art language models are capable of generating mixed-language utterances, but they often sound unnatural; improving the quality of future language models, such that they can generate bilingual utterances in a way that sounds natural to bilingual speakers, remains a challenge. To address this challenge, more linguistic work is required, focusing on identifying the types of "errors" (i.e., deviations from typical bilingual language) that are most off-putting to users, so that such errors can be avoided. Acknowledging that bilingual language in general and CS in particular often has "dialects", and that switches that sound natural to some users might be totally unacceptable to others (Balam et al., 2020), only highlights the importance of this challenge. Our work underscores the fact that if multilingual language technology fails to avoid unnatural and unacceptable grammatical patterns, it risks alienating users and even disrupting their ability to effectively complete the tasks that they wish to accomplish.

**References**

Aaron, J. E. (2015). Lone English-origin nouns in Spanish: The precedence of community

  norms. *International Journal of Bilingualism*, *19*(4), 459-480.

Ahn, E., Jimenez, C., Tsvetkov, Y., & Black, A. W. (2020, January). What code-switching

  strategies are effective in dialog systems? In A. Ettinger, G. Jarosz, & J. Pater (Eds.),

  *Proceedings of the society for computation in linguistics 2020* (pp. 254–264). Association

  for Computational Linguistics. https://aclanthology.org/2020.scil-1.32

Anderson, A. H., Bader, M., Bard, E. G., Boyle, E., Doherty, G., Garrod, S., et al. (1991). The

  HCRC map task corpus. *Language and Speech*, *34*(4), 351-366.

Anderson, A., Brown, G., Shillcock, R., & Yule, G. (1984). *Teaching talk: Strategies for

  production and assessment.* Cambridge University Press.

Auer, P. (1998). *Code-switching in conversation: Language, interaction and identity*. Routledge.

Balam, O., Parafita Couto, M. d. C., & Stadthagen-González, H. (2020). Bilingual verbs in three

  Spanish/English code-switching communities. *International Journal of Bilingualism*,

  *24*(5-6), 952– 967.

Beatty-Martínez, A. L., & Dussias, P. E. (2017). Bilingual experience shapes language

  processing: Evidence from codeswitching. *Journal of Memory and Language*, *95*, 173–

  189.

Beatty-Martínez, A. L., & Dussias, P. E. (2018). Tuning to languages: Experience-based

  approaches to the language science of bilingualism. *Linguistics Vanguard*, *4*(1).

Beatty-Martínez, A. L., & Dussias, P. E. (2019). Revisiting masculine and feminine grammatical

  gender in Spanish: Linguistic, psycholinguistic, and neurolinguistic evidence. *Frontiers

  in Psychology*, *10*, 751.

Beatty-Martínez, A. L., Navarro-Torres, C. A., & Dussias, P. E. (2020). Codeswitching: A bilingual toolkit for opportunistic speech planning. *Frontiers in Psychology*, *11*, 1699.

Beatty-Martínez, A. L., Navarro-Torres, C. A., Dussias, P. E., Bajo, M. T., Guzzardo Tamargo, R. E., & Kroll, J. F. (2020). Interactional context mediates the consequences of bilingualism for language and cognition. *Journal of Experimental Psychology: Learning, Memory, and Cognition*, *46*(6), 1022-1047.

Bell, A. (1984). Language style as audience design. *Language in Society*, *13*(2), 145–204. https://doi. org/10.1017/S004740450001037X

Bellamy, K., & Parafita Couto, M. C. (2022). Gender assignment in mixed noun phrases: State of the art. In Ayoun, D. (Ed.), *The Acquisition of gender: Crosslinguistic perspectives* (pp. 13-48). John Benjamins Publishing Company.

Berríos, J., Swain, A., & Fricke, M. (2023). Implementing the map task in applied linguistics research: What, how, and why. *Research Methods in Applied Linguistics*, *2*(3), 100081.

Bhatt, R. M. (1997). Code-switching, constraints, and optimal grammars. *Lingua*, *102*(4), 223–251. Chan, B. H.-s. (2009). English in Hong Kong Cantopop: Language choice, code-switching and genre. *World Englishes*, *28*(1), 107–129.

Blokzijl, J., Deuchar, M., & Parafita Couto, M. C. (2017). Determiner asymmetry in mixed nominal constructions: The role of grammatical factors in data from Miami and Nicaragua. *Languages*, *2*(4), 20.

Chan, B. H. S. (2009). English in Hong Kong Cantopop: language choice, code-switching and genre. *World Englishes*, *28*(1), 107-129.

Chang, C., Chuang, S., & Lee, H. (2018). Code-switching sentence generation by generative adversarial networks and its application to data augmentation. *CoRR*, *abs/1811.02356*.

http://arxiv.org/abs/ 1811.02356

Clegg, J. H. (2006). *Lone English-origin nouns in the Spanish of New Mexico: A variationist analysis of phonological and morphological adaptation* [Doctoral dissertation, University of New Mexico].

Clyne, M. G. (2003). *Dynamics of language contact: English and immigrant languages*. Cambridge University Press.

Deuchar, M. (2020). Code-switching in linguistics: A position paper. *Languages*, *5*(2), 22.

Fricke, M., & Kootstra, G. J. (2016). Primed codeswitching in spontaneous bilingual dialogue. *Journal of Memory and Language*, *91*, 181–201. https://doi.org/https://doi.org/10.1016/j.jml.2016.04.003

Gardner-Chloros, P., & Weston, D. (2015). Code-switching and multilingualism in literature. *Language and Literature*, *24*(3), 182–193.

Goldrick, M., Putnam, M., & Schwarz, L. (2016). Coactivation in bilingual grammars: A computational account of code mixing. *Bilingualism: Language and Cognition*, *19*(5), 857-876.

Gumperz, J. J. (1982). *Discourse strategies* (Vol. 1). Cambridge University Press.

He, H., Balakrishnan, A., Eric, M., & Liang, P. (2017, July). Learning symmetric collaborative dialogue agents with dynamic knowledge graph embeddings. In R. Barzilay & M.-Y. Kan (Eds.), *Proceed- ings of the 55th annual meeting of the association for computational linguistics (volume 1: Long papers)* (pp. 1766–1776). Association for Computational Linguistics. https://doi.org/10.18653/ v1/P17-1162

Honnibal, M., & Montani, I. (2017). *spaCy 2: Natural language understanding with Bloom embeddings, convolutional neural networks and incremental parsing* [To appear].

Kootstra, G. J., Dijkstra, T., & van Hell, J. G. (2020). Interactive alignment and lexical triggering of code-switching in bilingual dialogue. *Frontiers in Psychology*, *11*, 1747.

Kootstra, G. J., van Hell, J. G., & Dijkstra, T. (2010). Syntactic alignment and shared word order in code-switched sentence production: Evidence from bilingual monologue and dialogue. *Journal of Memory and Language*, *63*(2), 210-231.

Kootstra, G. J., van Hell, J. G., & Dijkstra, T. (2012). Priming of code-switches in sentences: The role of lexical repetition, cognates, and language proficiency. *Bilingualism: Language and Cognition*, *15*(4), 797-819.

Lenth, R. V. (2024). *Emmeans: Estimated marginal means, aka least-squares means* [R package version 1.10.4]. https://CRAN.R-project.org/package=emmeans

MacSwan, J. (2014). Programs and proposals in codeswitching research: Unconstraining theories of bilingual language mixing. *Grammatical theory and bilingual codeswitching*, 1-33.

McClure, E. (2001). Oral and written Assyrian–English codeswitching. In R. Jacobson (Ed.), *Codeswitching worldwide ii* (pp. 157–191, Vol. 126). Mouton de Gruyter Berlin.

Milroy, J., et al. (1995). *One speaker, two languages: Cross-disciplinary perspectives on code-switching*. Cambridge University Press.

Muysken, P. (2000). *Bilingual speech: A typology of code-mixing*. Cambridge: Cambridge University Press.

Myers-Scotton, C. (1993). Common and uncommon ground: Social and structural factors in codeswitching. *Language in Society*, *22*(4), 475–503.

Myers-Scotton, C. (2002). *Contact linguistics: Bilingual encounters and grammatical outcomes*. Oxford University Press, USA.

Myers-Scotton, C., & Gross, S. (2002). Making a minimalist approach to codeswitching work:

Adding the matrix language. *Bilingualism: Language and Cognition*, *5*, 69–91.

https://doi.org/10.1017/ S1366728902000147

Nenkova, A., Gravano, A., & Hirschberg, J. (2008, June). High frequency word entrainment in spoken dialogue. In J. D. Moore, S. Teufel, J. Allan, & S. Furui (Eds.), *Proceedings of ACL-08: Hlt, short papers* (pp. 169–172). Association for Computational Linguistics.

https://aclanthology.org/P08- 2043

OpenAI, Achiam, J., Adler, S., Agarwal, S., Ahmad, L., Akkaya, I., Aleman, F. L., Almeida, D., Al- tenschmidt, J., Altman, S., Anadkat, S., Avila, R., Babuschkin, I., Balaji, S., Balcom, V., Baltescu, P., Bao, H., Bavarian, M., Belgum, J., . . . Zoph, B. (2024). Gpt-4 technical report.

Otheguy, R., & Lapidus, N. (2003). An adaptive approach to noun gender in New York contact Spanish. *A Romance perspective on language knowledge and use*, 209–229.

Parekh, T., Ahn, E., Tsvetkov, Y., & Black, A. W. (2020, November). Understanding linguistic accommodation in code-switched human-machine dialogues. In R. Fernández & T. Linzen (Eds.), *Proceedings of the 24th conference on computational natural language learning* (pp. 565–577). Association for Computational Linguistics.

https://doi.org/10.18653/v1/2020.conll-1.46

Pfaff, C. W. (1979). Constraints on language mixing: Intrasentential code-switching and borrowing in Spanish/English. *Language*, *55*(2), 291–318.

Pickering, M., & Garrod, S. (2004). Toward a mechanistic psychology of dialogue. *The Behavioral and Brain Sciences*, *27*, 169–90.

https://doi.org/10.1017/S0140525X04000056

Poplack, S. (1980). Sometimes I'll start a sentence in Spanish y termino en Español: Toward a typology of code-switching. *Linguistics*, *18*(7-8), 581–618.

Reitter, D., & Moore, J. D. (2014). Alignment and task success in spoken dialogue. *Journal of Memory and Language*, *76*, 29-46.

Rijhwani, S., Sequiera, R., Choudhury, M., Bali, K., & Maddila, C. S. (2017). Estimating code-switching on Twitter with a novel generalized word-level language detection technique. *Proceedings of the 55th Annual Meeting of the Association for Computational Linguistics*, 1971–1982. https://doi.org/10.18653/v1/P17-1180

Rizvi, M. S. Z., Srinivasan, A., Ganu, T., Choudhury, M., & Sitaram, S. (2021, April). GCM: A toolkit for generating synthetic code-mixed text. In D. Gkatzia & D. Seddah (Eds.), *Proceedings of the 16th conference of the European chapter of the Association for Computational Linguistics: System demonstrations* (pp. 205–211). Association for Computational Linguistics. https://doi.org/10. 18653/v1/2021.eacl-demos.24

Salvador, S., & Chan, P. (2007). Toward accurate dynamic time warping in linear time and space. *Intelligent Data Analysis*, *11*(5), 561–580.

Samanta, B., Reddy, S., Jagirdar, H., Ganguly, N., & Chakrabarti, S. (2019). A deep generative model for code-switched text. *Proceedings of the Twenty-Eighth International Joint Conference on Artificial Intelligence*.

Sankoff, D., & Poplack, S. (1981). A formal grammar for code-switching. *Research on Language & Social Interaction*, *14*(1), 3–45.

Schwieter, J. W., & Festman, J. (2023). *The cognitive neuroscience of bilingualism*. Cambridge University Press.

Sebba, M., Mahootian, S., & Jonsson, C. (2012). *Language mixing and code-switching in writing: Approaches to mixed-language written discourse*. Routledge.

Thompson, H. S., Anderson, A., Bard, E. G., Doherty-Sneddon, G., Newlands, A., & Sotillo, C. (1993). The HCRC map task corpus: Natural dialogue for speech recognition. *Human Language Technology: Proceedings of a Workshop Held at Plainsboro, New Jersey, March 21-24, 1993*. https://aclanthology.org/H93-1005

Torres Cacoullos, R., & Travis, C. E. (2018). *Bilingualism in the community: Code-switching and grammars in contact*. Cambridge University Press.

Trofimovich, P., & Kennedy, S. (2014). Interactive alignment between bilingual interlocutors: Evidence from two information-exchange tasks. *Bilingualism: Language and Cognition*, *17*(4), 822–836.

Valdés Kroff, J. R. (2016). Mixed NPs in Spanish-English bilingual speech: Using a corpus-based approach to inform models of sentence processing. In R. E. Guzzardo Tamargo, C. M. Mazak, & M. C. Parafita Couto (Eds.), *Issues in Hispanic and Lusophone linguistics, vol. 11. Spanish- English codeswitching in the Caribbean and the US* (pp. 281–300). John Benjamins. https://doi.org/http://dx.doi.org/10.1075/ihll.11.12val

Vaswani, A., Shazeer, N., Parmar, N., Uszkoreit, J., Jones, L., Gomez, A. N., Kaiser, L., & Polosukhin, I. (2017). Attention is all you need. *Proceedings of the 31st Conference on Neural Information Processing Systems (NIPS 2017)*.

**Table 1.** *Overall summary of participant demographics and language history information. All self-ratings used a scale of 0 - 100. *Questions regarding birthplace and current residence were added partway through the Alternational Experiments. **Participants were first asked to enter their "native language, or the language they are providing responses for", followed by their "most proficient second language, or the language they are providing responses for", to account for the fact that many participants considered themselves simultaneous bilinguals. The questionnaire asked for age of acquisition (AoA) information only for the "most proficient second language", which is what we report in this row. There were a small number of missing responses or responses regarding a language other than English or Spanish.*

| | | Exp 1 (Alternational) | Exp 2 (Insertional) |
|---|---|---|---|
| n | | 218 (47% male) | 171 (45% male) |
| age | mean (SD) | 32.0 (10.6) | 32.1 (9.7) |
| highest level of education | less than high school | 7 (3%) | 1 (1%) |
| | high school | 21 (10%) | 19 (11%) |
| | trade school / other | 3 (1%) | 3 (2%) |
| | (some) college | 144 (66%) | 100 (58%) |
| | (some) graduate school | 43 (20%) | 48 (28%) |
| native language(s) | Eng only | 99 (45%) | 88 (51%) |
| | Eng + other | 2 (1%) | 0 |
| | Spa only | 59 (27%) | 41 (24%) |
| | Spa + other | 1 (< 1%) | 0 |
| | Eng + Spa | 57 (26%) | 41 (24%) |
| | Eng + Spa + other | 0 | 1 (1%) |
| Region of Birth * (n participants) | U.S. | 119 (77%) | 128 (75%) |
| | Mexico or Central America | 14 (9%) | 16 (9%) |
| | South America | 10 (6%) | 14 (8%) |
| | Caribbean | 4 (3%) | 4 (2%) |
| | Other (Europe, Asia, Africa) | 7 (5%) | 8 (5%) |
| | missing data | 64 | 1 |
| Current Residence * (n participants) | CA | 33 (22%) | 35 (21%) |
| | TX | 27 (18%) | 20 (12%) |
| | FL | 18 (12%) | 22 (13%) |
| | NY | 14 (9%) | 9 (5%) |
| | Midwest (IL, OH, IN, etc.) | 16 (11%) | 17 (10%) |
| | Other Northeast (NJ, MA, PA, etc.) | 16 (11%) | 21 (13%) |
| | Other West (WA, AZ, CO, etc.) | 14 (9%) | 13 (8%) |
| | Other South (NC, VA, GA, etc.) | 11 (7%) | 27 (16%) |
| | missing data | 69 | 7 |
| AoA for "most proficient second language" ** | for L2 Eng; mean (SD) [n responses] | 7.9 (7.5) [n=68] | 8.7 (9.1) [n=46] |
| | for L2 Spa; mean (SD) [n responses] | 5.5 (6.6) [n=143] | 5.9 (6.7) [n=114] |
| self-rated Eng proficiency | grand mean of speak, understand, read, write (SD) | 93.2 (14.2) | 96.2 (10.9) |
| self-rated Spa proficiency | grand mean of speak, understand, read, write (SD) | 77.4 (25.0) | 68.9 (22.2) |
| how likely to mix languages with friends? | mean (SD) | 53.0 (34.0) | 52.2 (33.5) |
| how likely to mix languages with family? | mean (SD) | 55.0 (36.4) | 61.6 (36.1) |
| how likely to mix languages at work? | mean (SD) | 41.7 (34.7) | 44.5 (34.1) |
| enjoy mixing languages? | mean (SD) | 59.2 (32.3) | 59.4 (30.0) |

**Table 2**

*Participants' self-reported language background in Experiment 1, broken down by condition.*

| CS Strategy | Total n Participants | Eng | Spa | Both | Eng Prof | Spa Prof |
|---|---|---|---|---|---|---|
| Baseline | 46 | 28 | 9 | 9 | 88.6 | 68.2 |
| Alignment | 43 | 25 | 6 | 12 | 92.9 | 67.5 |
| Adversarial | 43 | 19 | 15 | 9 | 90.0 | 71.2 |
| Random | 41 | 16 | 18 | 7 | 85.4 | 68.7 |
| Short-context | 45 | 13 | 12 | 20 | 88.3 | 67.3 |
| Total | 218 | 101 (46%) | 60 (28%) | 57 (26%) | 89.1 (sd=13.5) | 68.6 (sd=20.6) |

*Note.* *Eng* and *Spa* are the numbers of participants who listed each language as a native language. *Both* is the number of participants reporting both English and Spanish as native languages. Three participants reported an additional native language other than English or Spanish: there was one English-French, one English-Urdu, and one Spanish-Korean speaker. For simplicity, these participants are counted above as *Eng*, *Eng*, and *Spa*, respectively. *Eng Prof* and *Spa Prof* refer to self-rated mean proficiency in speaking, understanding, reading, and writing each language.

**Table 3.** *Detailed descriptive statistics for the dialogs collected in Experiments 1 and 2. Cells that differ from each other (for Task Success, Experiment 2) or from the baseline condition (all other cases) according to post-hoc t-tests are shaded and marked with asterisks (\* p < .05, \*\* p < .01, \*\*\* p < .001). For NP switches in Experiment 2, relative proportions are calculated within each condition, and cells whose absolute Pearson residuals are greater than 2.0 are marked with an asterisk (\*).*

| | | Experiment 1: Alternational Strategies | | | | | | Experiment 2: Insertional Strategies | | | | |
|---|---|---|---|---|---|---|---|---|---|---|---|---|
| | | Baseline | Alignment | Adversarial | Random | Short-Context | Average or Total | Baseline | Cong | Fem. Incong. | Masc. Incong. | Average or Total |
| Dialog Metrics | # Dialogs | 184 | 172 | 172 | 164 | 180 | 174.4 (6.9) | 176 | 164 | 168 | 176 | 171 (5.1) |
| | Mean # Utts/Dialog | 9.4 | 9.4 | 8.7 | 9.5 | 9.4 | 9.3 (3.6) | 9.0 | 9.2 | 9.6 | 9.3 | 9.3 (3.5) |
| | Mean # Tokens/Utt | 7.8 | 9.6 | 8.8 | 6.7 | 8.3 | 8.3 (10.1) | 8.0 | 6.7 | 8.4 | 6.7 | 7.4 (7.0) |
| | % Eng | 55.3 | 51.4 | 47.2 | 55.0 | 52.2 | 52.2 (36.3) | 18.2 | 14.1 | 15.1 | 23.5 | 17.8 (34.4) |
| | % Spa | 27.6 | 34.0 | 32.6 | 28.6 | 32.3 | 31.0 (32.7) | 71.5 | 65.3 | 62.9 | 61.2 | 65.2 (37.0) |
| | % Mixed | 8.2 | 8.1 | 16.3 | 8.5 | 9.7 | 10.2 (17.8) | 7.7 | 10.5 | 16.5 | 9.4 | 11.0 (18.0) |
| | % IS | 6.0 | 3.1 | 32.9\*\*\* | 20.6\*\*\* | 18.2\*\*\* | 16.0 (23.2) | 2.2 | 2.9 | 2.6 | 2.4 | 2.5 (9.0) |
| | % IS Entrainment | 84.4 | 89.0 | 66.5\*\*\* | 73.4\*\* | 67.4\*\*\* | 76.2 (26.0) | 80.1 | 81.9 | 82.9 | 75.8 | 80.1 (32.4) |
| Task Success Metrics | % Games Complete | 90 | 88 | 77 | 84 | 86 | 85 (36) | 72 | 66 | 80 | 68 | 71 (45) |
| | Game Time (sec.) | 268.1 | 282.5 | 301.4 | 284.4 | 281.0 | 283.2 (99.9) | 324 | 325 | 309\* | 322 | 320 (95) |
| | Route Distance | 2.8 | 2.7 | 2.4 | 3.1 | 2.5 | 2.7 (2.7) | 3.1 | 3.6 | 2.2\*\* | 3.1 | 3.0 (2.8) |
| | Task Enjoy (0-100) | 83.6 | 79.7 | 79.1 | 70.0\*\* | 77.7 | 78.2 (23.3) | 86.0 | 77.9\* | 84.1 | 74.7\*\* | 80.7 (21.2) |
| | Task Success (0-100) | 74.2 | 73.1 | 74.0 | 66.8 | 71.5 | 72.0 (23.6) | 69.5 | 65.1 | 75.0\*\* | 67.8\*\* | 69.0 (23.8) |
| | Diff. Comm. (0-100) | 29.8 | 30.0 | 36.5 | 43.5\*\* | 32.2 | 34.2 (29.7) | 37.3 | 42.3 | 31.7 | 48.6\* | 40.0 (29.4) |
| | Diff. Ins. (0-100) | 33.7 | 38.2 | 37.0 | 35.8 | 33.5 | 35.6 (29.1) | 36.7 | 38.2 | 25.7\* | 42.4 | 35.8 (27.8) |
| NP Switches | # Fem. Cong. NPs | 0 | 1 | 5 | 6 | 4 | 16 | 1 (0.06) | 14 (0.14) | 20 (0.18) | 13 (0.22) | 48 |
| | # Fem Incong. NPs | 5 | 10 | 5 | 4 | 13 | 37 | 5 (0.29) | 34 (0.33) | 43 (0.39) | 3\* (0.05) | 85 |
| | # Masc. Cong. NPs | 7 | 7 | 10 | 15 | 13 | 52 | 10 (0.59) | 53 (0.51) | 46 (0.42) | 35 (0.60) | 144 |
| | # Masc. Incong. NPs | 0 | 0 | 1 | 0 | 0 | 1 | 1 (0.06) | 2 (0.02) | 1 (0.01) | 7\* (0.12) | 11 |
| | Total # Mixed NPs | 12 | 18 | 21 | 25 | 30 | 106 | 17 | 103 | 110 | 58 | 288 |

**Table 4**

*Participants' self-reported language background in Experiment 2, broken down by condition.*

| CS Strategy | Total n Participants | Eng | Spa | Both | Eng Prof | Spa Prof |
|---|---|---|---|---|---|---|
| Baseline | 44 | 19 | 12 | 13 | 88.2 | 63.8 |
| Congruent | 41 | 25 | 6 | 10 | 92.4 | 59.7 |
| Fem. Incong. | 42 | 19 | 13 | 10 | 90.0 | 69.4 |
| Masc. Incong. | 44 | 25 | 10 | 9 | 90.4 | 67.0 |
| Total | 171 | 88 (51%) | 41 (24%) | 42 (25%) | 90.2 (sd=11) | 65.1 (sd=19) |

*Note. Eng* and *Spa* are the numbers of participants who listed each language as a native language. *Both* is the number of participants reporting both English and Spanish as native languages. One participant reported three native languages (English, Spanish, and Arabic), and this participant is included in the *Both* category above. *Eng Prof* and *Spa Prof* refer to self-rated mean proficiency in speaking, understanding, reading, and writing each language.

**Table 5**

*Comparison of the current experiments with the subset of the* Common Friends *experiments that yielded the highest success rates.*

| | CS Strategy | # Dial. | % Success | # Utts./Dial. | # Tokens/Utt. | % Mix |
|---|---|---|---|---|---|---|
| Ahn et al. (2020) | Spa Ins Eng (Informal) | 44 | 64 | 8.6 | 6.0 | 37 |
| | Eng Alt Spa (Informal) | 47 | 64 | 7.7 | 6.1 | 37 |
| | Eng Ins Spa (Informal) | 44 | 77 | 7.4 | 5.7 | 44 |
| Parekh et al. (2020) | Hin Ins Eng | 41 | 63 | 8.6 | 6.0 | 47 |
| | Eng Alt Hin | 39 | 54 | 9.3 | 5.8 | 46 |
| | Eng Ins Hin | 41 | 76 | 9.5 | 6.3 | 53 |
| Current | Alternational | 872 | 85 | 9.3 | 8.3 | 10 |
| | Insertional | 684 | 71 | 9.3 | 7.4 | 11 |

*Note.* # *Dial.* is the number of dialogues produced, *% Success* is the percentage of successful games, # *Utts./Dial.* is the average number of utterances per dialog, # *Tokens/Utt.* the average number of tokens per utterance, and *% Mix* is the percentage of mixed utterances.

**Figure 1**

*The four maps used in our experiments.*

(a) Map 1

(b) Map 2

(c) Map 3

(d) Map 4

**Figure 2**

*Illustration of the client-side interface for participants to play the roles of (a) instructor and (b) navigator.*

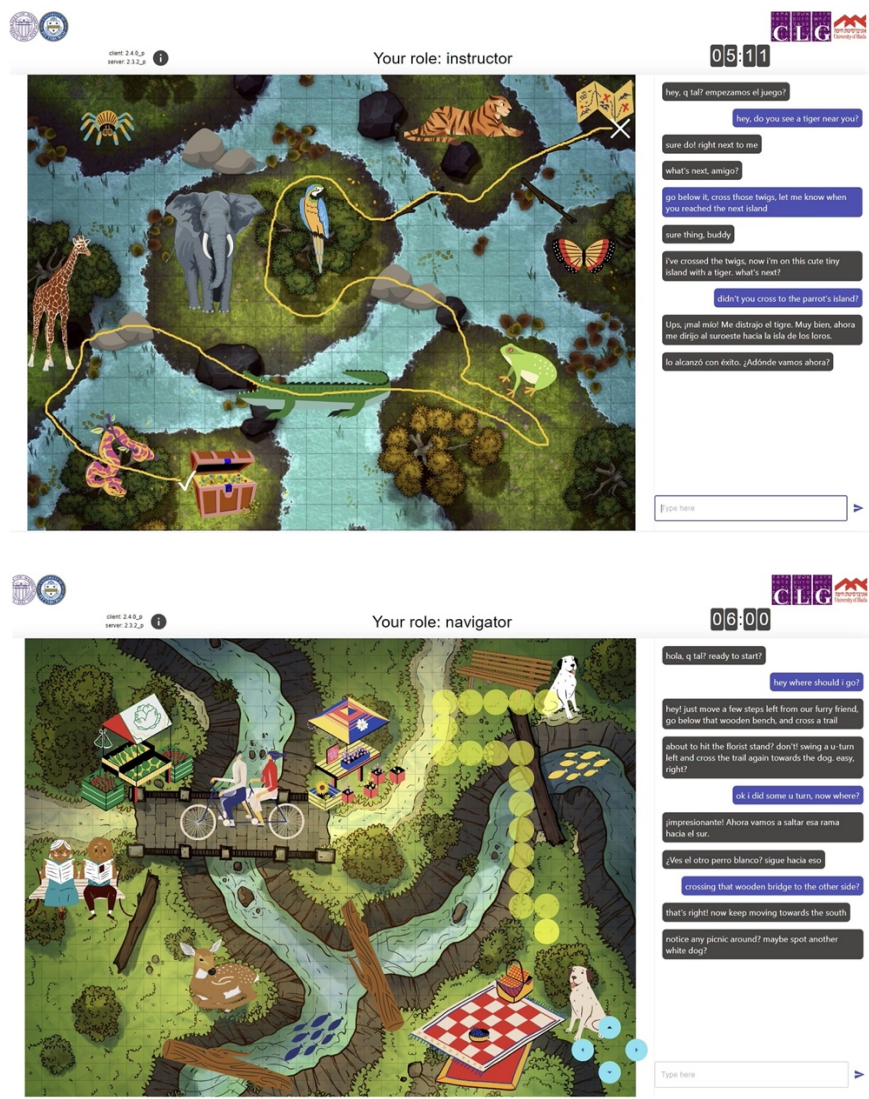

**Figure 3**

*Interaction between CS Strategy and Role in the analysis of Game Time in Experiment 2. The slowing effect of the navigator role was significant in all conditions except for the* Feminine Incongruent *strategy.*

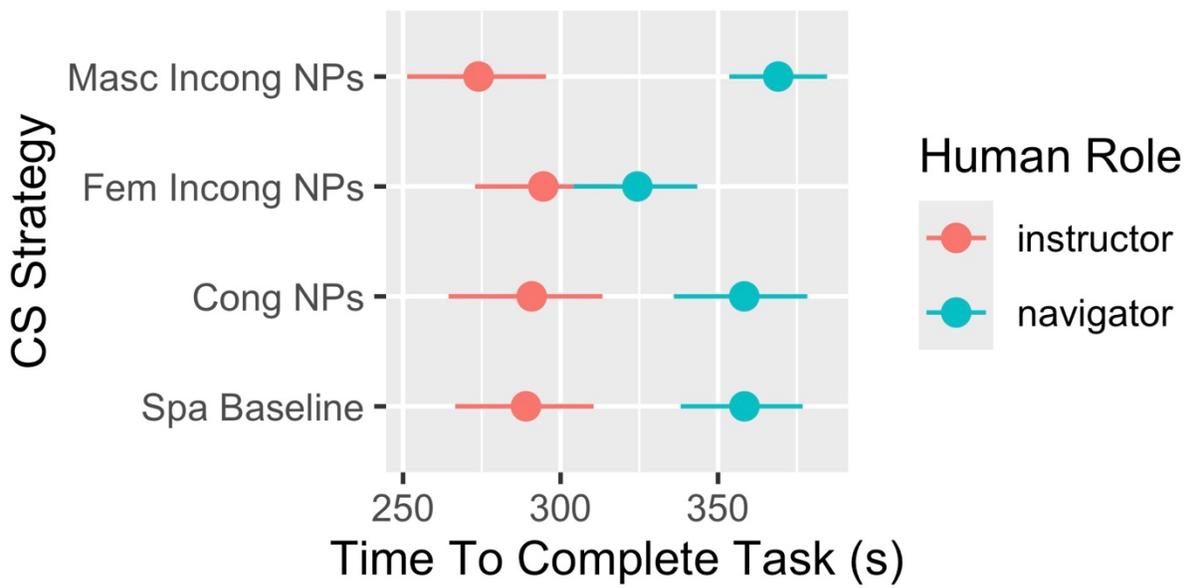